\documentclass[letterpaper, 10 pt, journal, twoside]{IEEEtran}

\author{Yue Hu, Shaoheng Fang, Weidi Xie and Siheng Chen 
\vspace{-4mm}
\thanks{Manuscript received: July 11, 2022; Revised November 19, 2022; Accepted January 16, 2023 .}
\thanks{\scriptsize{This paper was recommended for publication by Editor Pauline Pounds upon evaluation of the Associate Editor and Reviewers' comments. Yue Hu, Shaoheng Fang, Weidi Xie, and Siheng Chen are with Cooperative Medianet Innovation Center (CMIC), the School of Electronic Information and Electrical Engineering, Shanghai Jiao Tong University, Shanghai, 200240, China}
        {(e-mail:\tt 18671129361, shfang, weidi, sihengc@sjtu.edu.cn)}.
\scriptsize{Weidi Xie and Siheng Chen are also with Shanghai AI Laboratory. This research is partially supported by National Natural Science Foundation of China under Grant 62171276, the Science and Technology Commission of Shanghai Municipal under Grant 21511100900 and CALT Grant 2021-01.}}
\thanks{Digital Object Identifier (DOI): see top of this page.}
}

\title{Aerial Monocular 3D Object Detection}

\usepackage{graphicx}
\usepackage{amsmath}
\usepackage{amssymb}
\usepackage{booktabs}
\usepackage{ftnxtra}

\usepackage[pagebackref,breaklinks,colorlinks]{hyperref}
\usepackage[capitalize]{cleveref}
\crefname{section}{Sec.}{Secs.}
\Crefname{section}{Section}{Sections}
\Crefname{table}{Table}{Tables}
\crefname{table}{Tab.}{Tabs.}
\usepackage{multirow}
\usepackage[table,xcdraw]{xcolor}

\newcommand{\mypar}[1]{{\bf #1.}}

\begin{document}
\maketitle

\markboth{IEEE Robotics and Automation Letters. Preprint Version. Accepted January 16, 2023}
{Yue Hu \MakeLowercase{\textit{et al.}}: Aerial Monocular 3D Object Detection} 


\begin{abstract}
Drones equipped with cameras can significantly enhance human’s ability to perceive the world because of their remarkable maneuverability in 3D space. Ironically, object detection for drones has always been conducted in the 2D image space, which fundamentally limits their ability to understand 3D scenes. Furthermore, existing 3D object detection methods developed for autonomous driving cannot be directly applied to drones due to the lack of deformation modeling, which is essential for the distant aerial perspective with sensitive distortion and small objects. 
To fill the gap, this work proposes a dual-view detection system named \textbf{DVDET} to achieve aerial monocular object detection in both the 2D image space and the 3D physical space. To address the severe view deformation issue, 
we propose a novel trainable geo-deformable transformation module that can properly warp information from the drone's perspective to the birds' eye view (BEV). 
Compared to the monocular methods for cars, our transformation includes a learnable deformable network for explicitly revising the severe deviation. To address the dataset challenge, we propose a new large-scale simulation dataset named AM3D-Sim, 
and a new real-world aerial dataset named AM3D-Real
with high-quality annotations for 3D object detection.
Extensive experiments show that i) aerial monocular 3D object detection is feasible; ii) the model pre-trained on the simulation dataset helps real-world performance; and iii) DVDET also helps monocular 3D object detection for cars. To encourage more researchers to investigate this area, 
we released the \href{https://pan.baidu.com/s/1ZT9z4B5hvwJVFqwdEftkPQ?pwd=pdh3#list/path=%2F}{dataset} and \href{https://github.com/PhyllisH/DVDET}{related code}.
\end{abstract}
\begin{IEEEkeywords}
Aerial Systems: Perception and Autonomy
\end{IEEEkeywords}

\section{Introduction}
\label{sec:intro}
\IEEEPARstart{D}{rones} equipped with cameras have been actively used in a wide range of applications, including search and rescue, security and surveillance~\cite{SwarmSurvey3}. These real-world applications require 3D scene understanding. Drones' aerial perspective has higher flexibility and a larger perception range, leading to significant advantages in complex environments. However, the current drones' object detection is only limited to the 2D image space and the resulting 2D boxes with no 3D physical meaning~\cite{AerialDataset1}. LiDAR, providing precise 3D measurements~\cite{lang2019pointpillars} is not suitable for drones due to its heavy weight. Here, we consider the task of 3D object detection for aerial images.

\begin{figure}[!t]
  \centering
\includegraphics[width=0.95\linewidth]{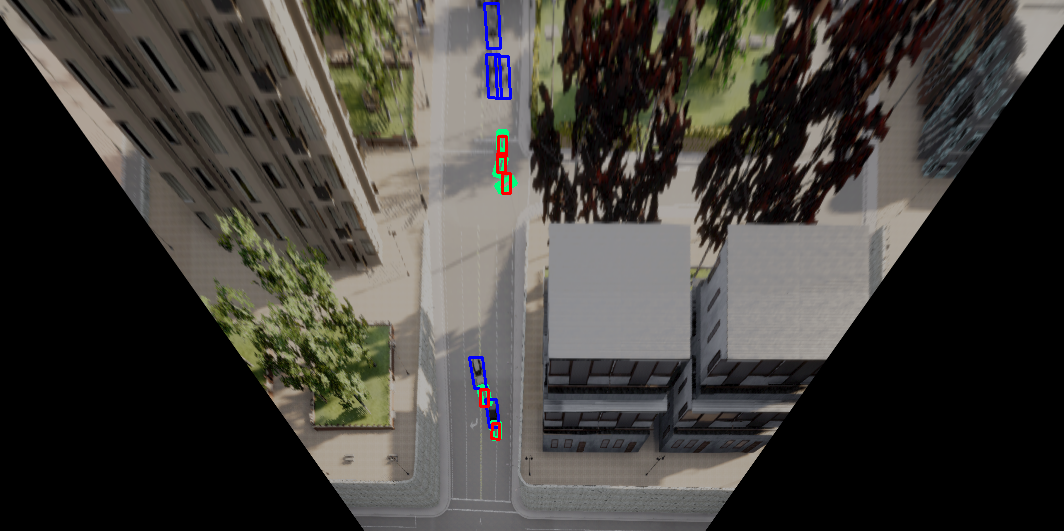}
\vspace{-2mm}
  \caption{Directly transforming 2D detection from RV to BEV fails due to the small object size and severe deformation issue. \textcolor{green}{Green}, \textcolor{blue}{blue}, and \textcolor{red}{red} is the ground truth, directly transformation of 2D detection, and detection of DVDET. Note that the BEV image is for reference only. Its pixel values are inaccurate due to deformation and loss of altitude information.}
  \label{fig:autonomous_method}
\vspace{-6mm}
\end{figure}

In practice, developing 3D object detection in aerial images faces three critical challenges: lack of a large-scale benchmark, a suitable 3D representation from drone's view, and a well-designed detection method. First, the existing drone's perception datasets~\cite{AerialDataset1,AerialDataset6} only have 2D annotations in image coordinate, which cannot be used for 3D object detection. Second, common 3D bounding box representation for autonomous driving is not suitable for drones, as in the aerial view, the object height is negligible compared to the flying height of the drones so that is almost impossible to estimate. 
Third, the intuitive method that directly transforms 2D boxes to 3D boxes fails in an aerial view due to the severe deformation issues caused by large aerial-view variations and distant imaging, see Fig.~\ref{fig:autonomous_method}. For the same reason, the emerging monocular 3D object detection methods for autonomous driving are not suitable for drones.

\begin{figure}[!t]
  \centering
\includegraphics[width=0.88\linewidth]{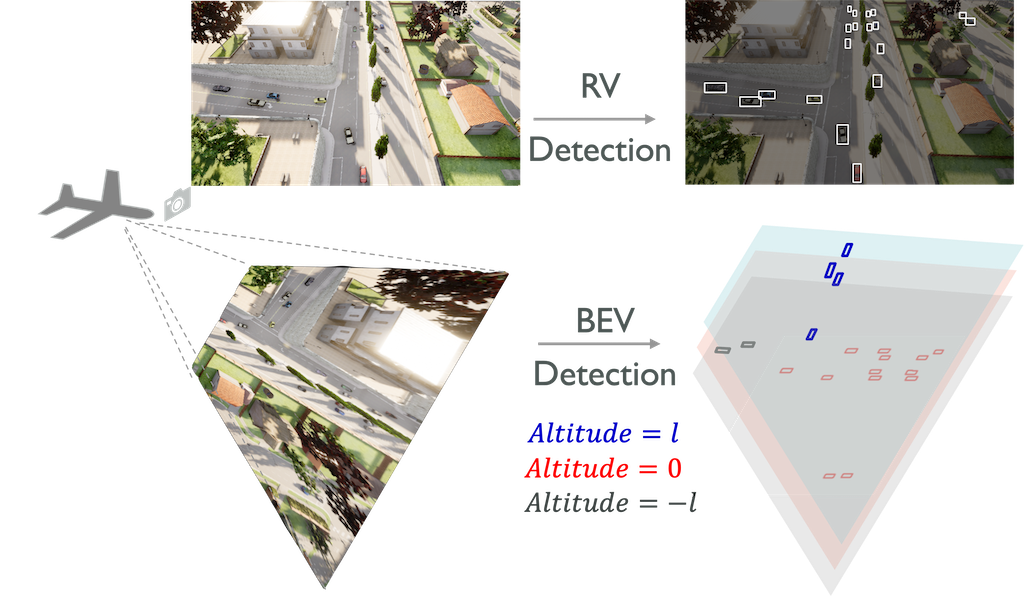}
\vspace{-4mm}
  \caption{Our dual-view object detection system simultaneously detects the objects in both 2D range view (RV) and 3D birds' eye view (BEV), given a 2D aerial image. Colors denote BEV detections at various altitudes, where 0 is the horizontal plane, and negatives are the planes below the horizontal.}
  \label{fig:intro}
\vspace{-7mm}
\end{figure}

To resolve the dataset limitation, we first propose a comprehensive and well-organized dataset, including simulation and real-world data versions: 
AM3D-Sim, AM3D-Real. 
AM3D-Sim is co-simulated by AirSIM~\cite{Airsim} and Carla~\cite{Carla}, where AirSIM simulates the flying drones and Carla simulates the complex background scenes and dynamic foreground objects. AM3D-Real is collected with DJI drones to validate the capability of the 3D measurement of the real world. 
Similar to 3D datasets for autonomous driving, 
our dataset contains aerial images with 2D/3D annotations.

To address the object representation issue, we propose a novel representation for drones: the BEV bounding box and the categorical altitude level, which simultaneously localizes an object on the ground and reflects its altitude. 
Categorical altitude estimation is essential in aerial view, 
as i) unlike cars, the observed ground from drones is usually much broader and non-flat, especially overpasses and ramps, so altitude is critical for 3D detection; ii) estimated altitudes allow to place 2D image information at the correct 3D locations, so altitude aids in precise warping. Note that we estimate the categorical altitude, instead of a continuous value. This can relieve the difficulty of altitude learning.

To address severe deformation issues caused by view variation and distant imaging in aerial view, which are rarely considered in autonomous driving, we propose a novel geo-deformable transformation, leveraging both the stability of the geometric transformation and the flexibility of the learnable deformable transformation. In the geometric transformation branch, the range-view (RV) feature is warped to the BEV based on camera pose information and the estimated altitude level. In the deformable transformation branch, we adjust the BEV features in a local region through a distance-aware deformable convolutional network (DCN). A residual structure combines the outputs of two branches. Compared to methods for 
autonomous driving, we additionally use a learnable DCN to adjust the geometrically warped feature, thus mitigating the severe deviation in the aerial view.

Based on the above designs, 
we propose a dual-view, aerial monocular 3D object detection system; termed as~\textbf{DVDET}, which jointly localizes the objects in the image and the 3D space.
We utilize the domain transfer technique to help handle the real-world data with the knowledge learned on the large-scale simulation dataset. We conduct comprehensive experiments to validate the effectiveness of DVDET.

To summarize, our contributions are as follows:

$\bullet$ We propose a novel task of aerial monocular 3D object detection to promote 3D scene understanding for drones from an aerial perspective. Our final system can simultaneously achieve 2D object detection in RV and 3D object detection in BEV given one 2D aerial image.

$\bullet$ We propose two core techniques specifically designed for aerial monocular 3D object detection, including trainable geo-deformable transformation, which warps the features from the RV to the BEV by leveraging camera pose parameters, geometric prior and learning ability, as well as categorical altitude estimation, which estimates the altitude level of each image pixel through classification.

$\bullet$ We build novel simulation and real-world benchmarks for the task of aerial monocular 3D object detection. We conduct extensive experiments to validate the proposed methods. We apply the domain transfer technique to benefit the real-world performance with the large-scale simulation data.


\section{Related Work}

\mypar{Aerial object detection}
Since objects in aerial perspective have massive variations in scale and orientation, existing detection datasets and algorithms could not be directly applied. To mitigate the dataset problem, a large number of aerial object detection datasets~\cite{AerialDataset1, AerialDataset6} with large quantities of arbitrarily oriented instances in complicated scenes are proposed. To handle the large scale and orientation variations along with aerial perspective, a large amount of aerial object detection algorithms are proposed, e.g. feature pyramids~\cite{lin2017feature} and deformable modules~\cite{dai2017deformable} are designed to handle the scale variations and the orientation variants. 

Unfortunately, the current aerial object detection datasets and algorithms only focus on the 2D range-view space and could not directly achieve 3D scene understanding. Recently, 3D object detection in driving scenarios is emerging~\cite{geiger2012Kitti,caesar2020nuscenes,sun2020waymo}. To take advantage of the maneuverability of drones and fill the huge scientific blank in aerial 3D object detection, this work firstly proposes a new task of aerial monocular 3D object detection and a well-organized dataset.

\mypar{Monocular 3D object detection}
Monocular 3D object detection aims to detect objects in the 3D space given the 2D image. It has three types: direct, depth-based, and grid-based. The direct methods first detect the 2D boxes and then use the geometric relation to regress the 2D boxes to 3D boxes~\cite{GDD}, which perform inferior without explicit depth information. The depth-based methods~\cite{3DDepth1} first estimate the depth map, which is combined with the image to generate the pseudo-3D point clouds. Then, 3D object detection could be performed. However, here depth estimation and detection are not trained in an end-to-end manner. The grid-based methods~\cite{3DGrid2} predict BEV grid representation and conduct the 3D detection on the grid. However, equal contribution of the image features along the projection ray causes repeated grid features. Recently,~\cite{reading2021categorical,philion2020lift} jointly performs depth estimation and detection, and uses the estimated depth to weight the contribution of image features to the grids.

The current monocular 3D object detection methods are designed for driving scenarios, mainly about depth learning. However, drones' aerial perspective encounters severe view variation and deformation. To tackle this issue, we introduce a novel geo-deformable transformation, leveraging geometric prior and learning ability, to achieve a more precise transformation from the range view to BEV.

\mypar{View transformation}
View transformation is a common module of many tasks, such as the view synthesis~\cite{rombach2021geometry}, multi-view pedestrian detection, and tracking~\cite{MultiviewX-dataset}. It has two types: geometric and parametric transformation. The non-parametric geometric transformation~\cite{MultiviewX-dataset} explicitly transforms the source view to the target view based on the camera projection, easy to deploy, and performs stably, while unable to estimate the unseen areas and tolerate the view deformation. The parametric transformation implicitly implements the view transformation with neural networks~\cite{li2021ViewTransformationLearnt} trained with GAN~\cite{rombach2021geometry}. It could infer unseen regions based on the context and flexibly adjust itself to cope with the deformation, but hard to fit the diverse and changing views.

Taking advantage of the stability of the geometric transformation and the flexibility of the learnable parametric transformation, we propose the hybrid geo-deformable transformation to address the view variation and distortion issue.


\begin{figure*}[t]
  \centering
  \includegraphics[width=0.95\linewidth]{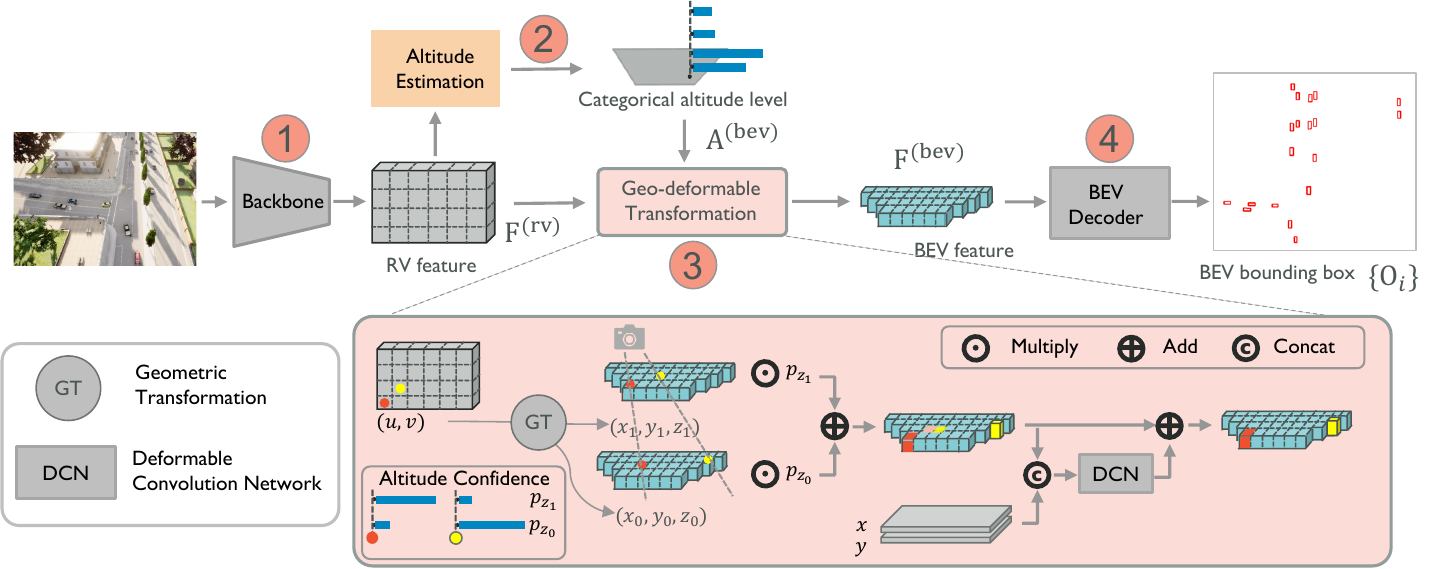}
  \vspace{-2mm}
  \caption{The overall framework of aerial monocular 3D object detection. First, a backbone is utilized to extract the RV feature from the image data. Second, the altitude estimation module predicts the categorical altitude level for each RV feature point, afterwards, a geometric transformation is performed to get the categorical altitude level for each coordinate in BEV. Third, the RV feature and the estimated altitudes are output to a geo-deformable transformation module to generate the BEV feature. Finally, the BEV feature is decoded to the object bounding boxes with orientation.}
  \label{fig:framework}
\vspace{-4mm}
\end{figure*}

\section{Problem Formulation}
\label{sec:Formulation}

Aerial monocular 3D object detection aims to localize the objects in the 3D space given a single 2D aerial image. To mathematically formulate this task, we represent a 3D object from an aerial perspective. Traditionally, an object's 3D bounding box is represented in two ways: eight box corners or box parameters, including the box center, the box size (length, width, height), and the box orientation. However, neither representation is suitable for the drone's cases. 
As the drone's altitude is usually tens or even hundreds of times higher than that of the objects, making it almost impossible to perceive the object's height or the precise altitude.

To address such a 3D object representation issue, we use a BEV bounding box with a categorical altitude to reflect the object's occupation on the ground and the discretized altitude level. Specifically, each object is parameterized by $(x,y,w,l,\theta,c,a)$, where $(x,y)$ is the center of the object on the ground, $w$/$l$/$\theta$ is the width/length/azimuth angle, $c$/$a$ is the object/altitude category. The altitude is categorized into 9 bins centered at $[-1.0, -0.5, 0, 0.5, 0.75, 1.0, 1.5, 2.0, 8.0]$, where 0 denotes the horizontal plane. The spacing is: a) densely and uniformly between -1m and 2m, including most objects; b) sparsely and non-uniformly between 2m to 8m, including most backgrounds with large altitude variance.

Since the aerial view is significantly different from car driving, most existing monocular 3D object detection methods designed for cars cannot be directly applied to drones. To fill this gap, we propose categorical altitude estimation and geo-deformable transformation in Section~\ref{sec:Method}. We further propose an overall perception system that jointly localizes objects in the 2D image and 3D space; see Sec.~\ref{System}.
\section{Dataset Preparation}
\label{Dataset}
To enable aerial monocular 3D object detection,  we develop the first datasets for 3D object detection for aerial image. The previous aerial object detection datasets only provide the 2D bounding boxes in the image coordinate system. Besides, the existing 3D object detection datasets for autonomous driving could not be easily transferred to drones due to the large domain gap, for example, perspectives. 

We propose both simulation and real-world datasets, named AM3D-Sim and AM3D-Real. The datasets include RGB images with well-annotated 2D \& 3D bounding boxes of vehicles and precise camera pose information; see Tab.~\ref{tab:dataset}. Our dataset organization refers to the database schema of NuScenes\cite{caesar2020nuscenes}, an open-source autonomous driving benchmark. Our goal is to motivate monocular 3D object detection from aerial view by providing challenging benchmarks with novel difficulties to the 3D perception community.

AM3D-Sim is collected by the co-simulation of CARLA and AirSIM. CARLA~\cite{Carla} simulates complex scenes and traffic flow, and AirSIM~\cite{Airsim} simulates drones flying in the scene. To promote data diversity, the flying height is set ranging from 40m to 80m, covering an area of 200m$\times$200m. In the simulation, the annotations could be produced autonomously, so we provide a large and diverse simulation benchmark. 


AM3D-Real is collected with DJI Matrice 300 RTK flying over the campus. The drone is equipped with a well-aligned LiDAR and an RGB camera. We annotate the 3D bounding boxes in the 3D point clouds collected by LiDAR and get the 2D boxes by projecting the 3D boxes back to the image based on the projection matrix, given by DJI SDK. Due to challenging and costly data collecting and labeling, the flying height is set lower and the dataset size is relatively smaller. 


\begin{table}[]
\vspace{-2mm}
\caption{Dataset statistics. FH is flying height in meter. *(*/*) denotes total(train/test).}
\vspace{-4mm}
\label{tab:dataset}
\begin{scriptsize}
\begin{tabular}{lcccc}
\hline
\multicolumn{1}{c}{Dataset} & Scenes & FH(m) & Images                & Boxes                    \\ \hline
AM3D-Sim                    & 3      & 40-80            & 48,250 (41,500/6,750) & 397,984 (347,588/5,0396) \\
AM3D-Real                   & 10     & 30-40            & 1,012 (919/93)        & 33,083 (31,668/1,415)    \\ \hline
\end{tabular}
\end{scriptsize}
\vspace{-7mm}
\end{table}

\vspace{-1mm}
\section{Methodology}
\vspace{-1mm}
\label{sec:Method}

\subsection{Motivation}
The aerial setting has more severe view deformation than the driving setting for two reasons. First, in autonomous driving, 
objects in the scene share almost the same altitude with the camera and the distortion along the height axis is less sensitive, so the ratio of height to depth is mostly constant.
The object height could be leveraged as a reliable proxy to estimate the depth information; while in the aerial view, there is no reliable proxy and we have to work with a pure 3D space problem. Second, autonomous driving usually considers detecting objects within 100 meters; while a drone usually considers more distant objects.

To tackle the 3D space problem in the aerial setting, we consider solutions from two aspects: i) achieve better altitude estimation; ii) compensate for the inaccurate altitude estimation. Accordingly, we propose two techniques: i) categorical altitude estimation, simplifies the challenging altitude learning by categorizing the altitude into multiple bins and substituting the strict regression task with an easygoing classification task; ii) geo-deformable transformation, corrects severe view deviation using a learnable deformable network with trainable offsets to compensate for the geometric spatial sampling generated based on the estimated altitudes.

\vspace{-4mm}
\subsection{Mathematical Framework}
As discussed in Sec.~\ref{sec:Formulation}, the goal is to provide the bird’s-eye-view (BEV) bounding box and the categorical altitude for each object in a range-view (RV) aerial image. Fig.~\ref{fig:framework} illustrates the proposed framework of monocular 3D object detection for drones in four steps. 

First, we extract the RV features from the RGB image by a backbone network. Given an image $\mathbf{I}\in \mathbb{R}^{ H_I \times W_I \times 3}$ with $H_I,W_I$ the image height and weight, the RV feature map is 

\begin{equation}
\mathbf{F}^{\rm (rv)} = f_{\rm backbone}(\mathbf{I}) \in \mathbb{R}^{H_R \times W_R \times C},
\vspace{-1mm}
\end{equation}
where $f_{\rm backbone}(\cdot)$ is a DLA-based backbone~\cite{yu2018dla}, and $H_R,W_R, C$ are the height, weight and channel dimension.

Second, we estimate the categorical altitude level for each coordinate in BEV by leveraging the proposed altitude estimation module followed by the geometric transformation. The BEV map of categorical altitude $\mathbf{A}^{\rm (bev)}$ is 
\begin{eqnarray}
\label{eq:altitude}
\mathbf{A}^{\rm (bev)}  & = & 
f_{\rm altitude}( \mathbf{F}^{\rm (rv)}  )
\in \mathbb{R}^ {X \times Y \times Z},
\vspace{-1mm}
\end{eqnarray}
where $f_{\rm altitude}(\cdot)$ is the proposed altitude estimation module, $X,Y$ denote the length of the perception field along $x$-axis and $y$-axis, $Z$ is the amount of categorical altitude bins along $z$-axis. Each element  $\mathbf{A}^{\rm (bev)}{(x,y,z)}$ reflects the confidence score of the $z$th altitude level at the location $(x,y)$ in BEV.

Third, we get the BEV features by warping the RV features based on the proposed geo-deformable transformation and the estimated categorical altitude. The BEV feature map is
\begin{equation}
\label{eq:transformation}
\mathbf{F}^{\rm (bev)} = f_{\rm deform}(\mathbf{F}^{\rm (rv)}, \mathbf{A}^{\rm (bev)}) \in \mathbb{R}^{ X \times Y\times C},
\vspace{-1mm}
\end{equation}
where $f_{\rm deform}(\cdot)$ is the proposed geo-deformable transformation module that leverages the camera pose information, geometric prior and the learning ability.

Fourth, we detect the BEV bounding boxes by decoding the BEV features. The detected BEV bounding boxes are

\begin{equation}
\{ \mathbf{o}_i \} \ = \  f_{\rm decoder}(\mathbf{F}^{\rm (bev)}, \mathbf{A}^{\rm (bev)})),
\end{equation}
\vspace{-1mm}
where $\mathbf{o}_i = (x_i,y_i,w_i,l_i,\theta_i, a_i, c_i)$ is the $i$th box with $(x_i,y_i)$  the center, $w_i$ the width, $l_i$  the length, $\theta_i$ the azimuth angle, $a_i$ the altitude level and $c_i$ the object category. Our decoder follows the established detector CenterNet~\cite{zhou2019objects}.

\vspace{-4mm}
\subsection{Categorical altitude estimation}
\label{sec:Altitude}
Here we elaborate on the details of the proposed categorical altitude estimation module in Equation~\eqref{eq:altitude}. The specific goal is to localize the object in $z$-axis. 
Categorical altitude estimation is critical for 3D detection in aerial view as the observed ground is broader and not flat, and allows to place 2D image information at the correct 3D locations, providing side information for precise warping. It is challenging because the aerial perspective encounters severe long-range issues, where the altitude difference across objects is relatively minor compared to the distance between objects and the drone. 
To tackle this issue, we categorize the altitude into multiple bins and substitute the regression task with a classification task.

To realize this, 
we first estimate the altitude level at each RV pixel. Intuitively, image context can provide rich altitude hints. For example, since most vehicles have similar sizes, they look bigger when they are closer to the drone and smaller when far away from the drone. The RV map of altitude confidence is obtained as
\begin{equation}
\mathbf{A}^{\rm (rv)} =  {\rm conv}( \mathbf{F}^{\rm (rv)}  )
\in \mathbb{R}^ {H_R \times W_R \times Z},
\end{equation}
where ${\rm conv}(\cdot)$ is a $1 \times 1$ convolution layer. Each element in $\mathbf{A}^{\rm (rv)}$ reflects the probabilities over all the altitude levels at each pixel in the RV.
We next geometrically transform the RV map of altitude confidence to the BEV coordinate by
\begin{equation}
\mathbf{A}^{\rm (bev)} =  
t (\mathbf{A}^{\rm (rv)} ) 
\in \mathbb{R}^ {X \times Y \times Z},
\end{equation}
where the geometric transformation $t(\cdot)$ is fully derived from the camera pose provided with the input image. The output $\mathbf{A}^{\rm (bev)}$ reflects the altitude information at each BEV location.

\vspace{-2mm}
\subsection{Geo-deformable transformation}
\label{Transformation}
The proposed geo-deformable transformation aims to learn the BEV feature for the 3D object detection given the RV feature; see Equation~\eqref{eq:transformation}. 
This is required to infer the 3D scene given the planar 2D image; however, the altitude information is not available without an extra depth sensor. 
Theoretically, each image pixel is the projection of a line across all the altitudes in the 3D space, 
causing ambiguities.

To compensate for the missing altitude information, 
we consider solutions from two aspects: 
1) weighting features along the $z$ axis; 
2) deforming features along with the $x,y$ axes. 
First, we leverage the geometric transformation derived from the camera pose to generate BEV representations at all possible altitudes; 
and then, weight them with the estimated altitude confidence. The weighted BEV representations are averaged along the altitude axis, collapsing to a flat BEV feature. 
Second, we use a trainable deformable convolutional network (DCN) to adaptively revise the distortion in the BEV feature caused by the imprecise altitude estimation, promoting flexibility in this view transformation phase. 
DCN is capable of augmenting the spatial sampling locations with additional offsets, which could help to finetune the geometrically transformed feature. 
A residual structure is applied to combine information from both the geometric transformation and the adaptively deformable transformation. 

\mypar{Geometric transformation} The geometric transformation is a non-parametric approach for view transformation. The camera projection matrix $\mathbf{P}$ defines the mapping between the global coordinate $(x,y,z)\in \mathbb{R}^3$ to the local image pixel coordinate $(u,v) \in \mathbb{R}^2$. The geometric transformation of altitude $z$ could be denoted as the following mapping:
\begin{equation}
    z\left[
    \begin{array}{c}
         u \\
         v \\
         1
    \end{array}
    \right]
    =
    \mathbf{P}
    \left[
    \begin{array}{c}
         x \\
         y \\
         z
    \end{array}
    \right].
\end{equation}

The RV feature $\mathbf{F}^{\rm (rv)} \in \mathbb{R}^{H_R \times W_R \times C}$ is transformed to all the $Z$ possible altitudes, which are stacked along the $z$-axis and produces the 3D feature $\mathbf{G}^{\rm (bev)} \in \mathbb{R}^{X \times Y \times Z \times C}$. Given the feature $\mathbf{G}$ and the altitude confidence $\mathbf{A}^{\rm (bev)}$ in BEV, via weighting and collapsing along the altitude, 
we can get the flattened BEV feature $\mathbf{F}^{\rm (bev)}_g\in \mathbb{R}^ {X \times Y \times C}$ as follows,
\begin{equation}
\begin{scriptsize}
    \mathbf{F}^{\rm (bev)}_g(x,y) =\frac{1}{Z}\sum_{z=0,1,...,Z-1} \mathbf{G}^{\rm (bev)}(x,y,z) \cdot \mathbf{A}^{\rm (bev)}(x,y,z),
\end{scriptsize}
\end{equation}
where $z$ is the index of the altitude level, $\mathbf{G}^{\rm (bev)}(x,y,z)$ is possible feature representation at the coordinate $(x,y,z)$. $\mathbf{A}^{\rm (bev)}(x,y,z)$ is the confidence that the altitude value of the feature point at the coordinate $(x,y)$ ranges in the $z$-th altitude level. Here, the 3D feature $\mathbf{G}^{\rm (bev)}$ contains the BEV feature across all the possible altitudes. As stated in~\cite{lang2019pointpillars}, BEV grids greatly reduce the computational overhead while offering similar performance to 3D voxel grids. So we use the flattened BEV feature while keeping the relative importance in the altitude levels to perform 3D object detection.

\mypar{Deformable transformation} 
Through geometric transformation across all the possible altitudes, we get flat yet stereo-like BEV feature. The non-parametric geometric transformation lacks learnable flexibility. Ideally, if the right altitude is localized, the BEV feature can exactly represent the real world. However, the severe long-range issue along with the aerial view makes the altitude estimation especially difficult. Therefore, the geometrically transformed BEV feature $\mathbf{F}^{\rm (bev)}_g$ is expected to encounter spatial sampling noise. 

To promote better transformation, 
a DCN layer is cascaded to augment the geometric spatial sampling with trainable offsets. 
We further concatenate the coordinates with the BEV feature to guide offset learning. Since the coordinates could hint at the network with the geometric prior that the perception field is increasing with the distance between objects and the camera, which means that the perturbation area is large at far distance, and vice versa.
Mathematically, the BEV feature map after the deformable convolution is
\begin{equation}
    \mathbf{F}^{\rm (bev)}_{d} \ = \ {\rm DCN}([\mathbf{F}^{\rm (bev)}_{g};\mathbf{X};\mathbf{Y}]),
\end{equation}
where ${\rm DCN}(\cdot)$ is the trainable DCN layer, 
$[;]$ denotes concatenation, $\mathbf{X},\mathbf{Y}\in \mathbb{R}^{X \times Y \times 1}$ refer to the $x,y$ coordinates of the feature points, 
reflecting the geometric prior.

Finally, we use a residual structure that combines the geometrically transformed feature $\mathbf{F}^{\rm (bev)}_{g}$ and the adaptively deformable feature $\mathbf{F}^{\rm (bev)}_{d}$ to get the final BEV feature map,
$
    \mathbf{F}^{\rm (bev)}=\mathbf{F}^{\rm (bev)}_{g}+\mathbf{F}^{\rm (bev)}_{d}.
$

\begin{figure}[!t]
  \centering
  \includegraphics[width=1.0\linewidth]{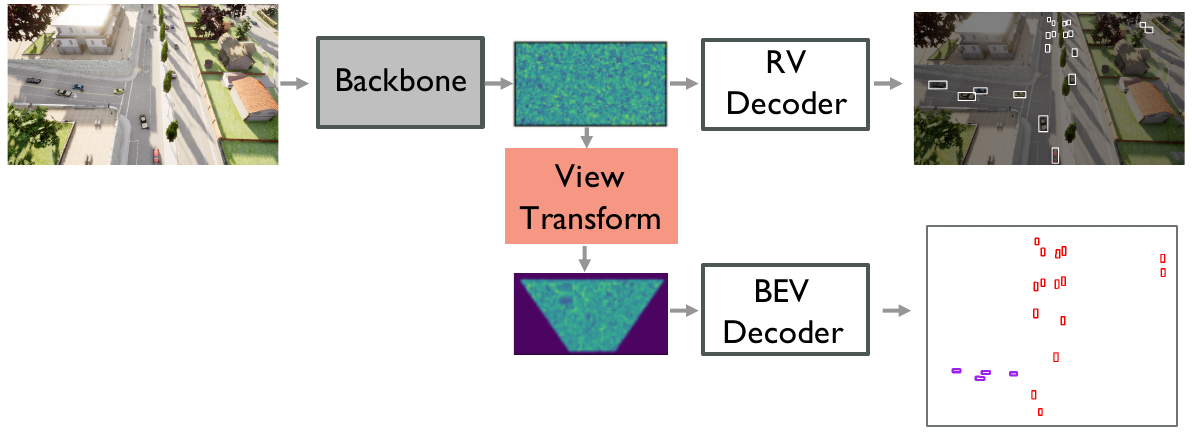}
    \vspace{-8mm}
  \caption{DVDET simultaneously localizes objects in image and 3D space. }
    \vspace{-6mm}
  \label{fig:system}
\end{figure}

\vspace{-2mm}
\subsection{Loss function}
The system considers two tasks: 
the altitude-level classification and the BEV-based object detection. 
For altitude-level classification, let $\mathbf{A}^{(\rm bev)}$ be the estimated altitude category, $\widetilde{\mathbf{A}}^{\rm (bev)}$ is the ground-truth altitude category, $\mathbf{M}$ is the objectiveness mask, 
only the foreground objects are supervised, the classification loss is then
$
    L_{\rm altitude} = \mathbf{M} \odot {\rm Focal}(
    \mathbf{A}^{\rm (bev)}, \widetilde{\mathbf{A}}^{\rm (bev)}).
$
Focal loss~\cite{lin2017focal} is used to alleviate the class imbalance issues. 
For BEV object detection, we follow CenterNet~\cite{zhou2019objects} and jointly optimize the classification and regression losses. 
Let $\mathbf{H}$ be the estimated category heatmap, 
$\widetilde{\mathbf{H}}$ be the ground-truth heatmap, 
the classification loss is $L_{\rm cls}={\rm Focal}(\mathbf{H}, \widetilde{\mathbf{H}})$. Let $(x,y,w,l,\theta)$ be a detected box and $(\widetilde{x},\widetilde{y},\widetilde{w},\widetilde{l},\widetilde{\theta})$ be the ground-truth box, $\ell_1$ loss $\left\| \cdot \right\|_1$ is used for the regression loss
\begin{eqnarray*}
    L_{\rm box} & = & \left\| x - \widetilde{x} \right\|_1 + \left\| y - \widetilde{y} \right\|_1 + \left\| w - \widetilde{w} \right\|_1 + \left\| l - \widetilde{l} \right\|_1 
    \\
    && + \left\| \sin \theta  -  \sin \widetilde{\theta} \right\|_1 + \left\| \cos \theta  -  \cos \widetilde{\theta} \right\|_1.
\end{eqnarray*}
The overall loss is the sum of $L_{\rm altitude}$, $L_{\rm cls}$ and $L_{\rm box}$. For dual-view system, the RV detection loss follows CenterNet~\cite{zhou2019objects} and is additionally summed to the final loss.

\begin{table*}[!ht]
\centering
\vspace{-1mm}
\caption{Overall performance of the baselines and our proposed geo-deformable transformation (\textit{GeoDT}) and categorical altitude estimation (\textit{CAE}) modules on AM3D-Sim. The BEV object detection is evaluated with AP/AP@50/AP@75. The categorical altitude estimation is measured with accuracy. Both \textit{GeoDT} and \textit{CAE} improves the performances.}
\vspace{-2mm}
\label{tab:MainResults}
\begin{scriptsize}
\begin{tabular}{l|cccccc|cccc|cccc}
\hline
\multicolumn{1}{c|}{\multirow{2}{*}{Method}} & \multicolumn{3}{c|}{\begin{tabular}[c]{@{}c@{}}Trans-\\ Location\end{tabular}} & \multicolumn{2}{c|}{\begin{tabular}[c]{@{}c@{}}Trans-\\ Type\end{tabular}} & \begin{tabular}[c]{@{}c@{}}Altitude\\ Estimate\end{tabular} & \multicolumn{4}{c|}{BEV Object Detection} & \multicolumn{4}{c}{\begin{tabular}[c]{@{}c@{}}Altitude Classification\end{tabular}} \\
\multicolumn{1}{c|}{}    & Early & Inter & \multicolumn{1}{c|}{Late}  & Geo   & \multicolumn{1}{c|}{Def} & CAE  & Fullset   & Town1  & Town2 & Town3    &Fullset  & Town1  & Town2  & Town3                 \\ \hline
\textit{Late-GeoT}         & - & - & \checkmark    & \checkmark  & -  & -                      &0.62/2.50/0      &1.13/5.89/0.05    & 0.37/2.38/0  & 0.07/0.48/0   & -      & -      & -   & -                     \\
\textit{Early-GeoT}        & \checkmark  & -  & -  & \checkmark  & -  & -                      &37.36/81.75/27.88  & 37.43/78.49/30.71   & 42.62/90.11/33.40    & 32.27/75.84/20.72    & -      & -      & -   & -                     \\
\textit{Inter-GeoT}        & -  & \checkmark  & -  & \checkmark  & -  & -                      &38.64/80.52/32.19  & 38.85/77.06/34.94   & 43.19/88.20/36.58    &34.17/75.42/26.15    & -      & -      & -   & -                     \\
\textit{Inter-GeoDT}       & -  & \checkmark  & -  & \checkmark  & \checkmark  &-              &40.84/81.85/36.37  & 43.39/80.32/42.67    &44.11/88.32/38.46   &36.02/76.79/30.05    & -      & -      & -   & -                     \\
\textit{Inter-GeoT-CAE}    & -  & \checkmark  & -  & \checkmark  & -  & \checkmark             &41.54/82.94/37.31  &43.27/80.56/42.35    &44.37/88.11/39.26    &37.65/79.61/31.96    & 88.56      & 74.55      & 93.54   & 83.54                    \\ 
\textit{DVDET}   & -  & \checkmark  & -  & \checkmark  & \checkmark  &\checkmark     &\textbf{42.70}/\textbf{84.57}/\textbf{38.37}     &\textbf{45.31}/\textbf{82.94}/\textbf{44.93}    & \textbf{45.06}/\textbf{89.43}/\textbf{39.41}    &\textbf{38.78}/\textbf{81.07}/\textbf{33.17}    & \textbf{90.36}     &\textbf{79.02}      & \textbf{94.24}   & \textbf{86.36}                    \\ \hline
\end{tabular}
\end{scriptsize}
\vspace{-4mm}
\end{table*}

\subsection{Simulation to real-world transfer}
To alleviate expensive and laborious real-world data collection/annotation,
we adopt a Sim2Real training regime, 
where the model is pre-trained on simulation data~(AM3D-Sim), 
and followed by fine-tuning on real data~(AM3D-Real).
Specifically, to minimize the domain gap, 
which can be triggered by the inconsistency between physical parameters,
{\em e.g.}~illumination, reflection, etc, 
we take inspiration from \textbf{domain randomization}~\cite{Tobin2017DomainRF}, conduct aggressive color augmentations on the simulation data, and train the model to be invariant to them.
Likely this model can adapt to the real-world environment, 
as the real visual scene is expected to be one sample in that rich distribution of training variations.


\section{Dual-View Object Detection System}
\label{System}
We further propose a dual-view object detection system, 
termed as \textit{DVDET}, 
which simultaneously perceive the objects in the 2D image space and the 3D physical space, based on the intuition that the two views could potentially promote each other. 
Specifically, the 2D image space can provide object details, 
such as color and shape, and help object understanding,
while the 3D space can provide more accurate spatial information. 
The implicit consistency from the supervision of the two views, 
including RV and BEV, can thus help reduce the error of each other 
and promote more precise detection. Fig.~\ref{fig:system} illustrates \textit{DVDET}. 
The detectors for two views share the same backbone. 
The RV decoder localizes objects in the 2D image space and the BEV decoder localizes objects in the 3D space by using the proposed categorical altitude estimation and geo-deformable transformation methods.


\section{Experimental Results} 

\subsection{Implementation details}
Our detector follows the CenterNet~\cite{zhou2019objects} with DLA-34~\cite{yu2018dla} backbone. The RV aerial image size is $(800, 450)$ and $(720,480)$ in the simulation and real-world dataset. The resolution of the BEV is $0.25$m/pixel. We transform the pyramid RV feature maps to BEV. The size of the BEV feature map is $(192,352)$ and $(96,128)$ in the simulation and real-world datasets respectively. We employ the generic detection evaluation metric: Average Precision (AP) at Intersection-over-Union (IoU) thresholds of 0.5 and 0.75.


\subsection{Overall performance}
\label{sec:overall}

\mypar{Evaluation on AM3D-Sim} 
We first compare baselines that directly transform the three different information sources:
detected objects, input image, 
or intermediate feature from RV to BEV according to the given camera pose information, as shown in Tab.~\ref{tab:MainResults}.  \textit{Early-GeoT}, \textit{Inter-GeoT} and \textit{Late-GeoT} denotes  transforming the raw RV image, intermediate feature and the detection output, respectively. Here~\textit{GeoT} is short for geometric transformation, 
fully derived from the camera pose. 
We see that: i) \textit{Inter-GeoT} performs the best;
ii) \textit{Early-GeoT} performs slightly inferior to \textit{Inter-GeoT}. 
Since compared to \textit{Inter-GeoT}, the input of \textit{Early-GeoT} has higher resolution, 
however, it encounters irreparable deformation, 
which could be alleviated in \textit{Inter-GeoT}. 
Overall, \textit{Inter-GeoT} has more flexibility and performs the best. 
iii) \textit{Late-GeoT} performs poorly. 
Directly transforming the RV detection fails the BEV object detection. 
This might be caused by two reasons: 
first, objects in the RV are mostly represented with an axis-aligned bounding box, which could not precisely represent the objects in the local coordinate; 
second, there are many tiny objects from the aerial perspective, occupying only 0.146\% of the image on average.
So, the view deformation along with the aerial perspective severely degrades the transformed BEV object detection performance.

\begin{table}[!t]
\centering
\vspace{-1mm}
\caption{Performance of dual-view system on AM3D-Sim. Dual-view system outperforms the individual views.}
\vspace{-2mm}
\label{tab:DualView-Sim}
\begin{small}
\begin{tabular}{l|ccc|ccc}
\hline
\multicolumn{1}{c|}{\multirow{2}{*}{System}} & \multicolumn{3}{c|}{BEV} & \multicolumn{3}{c}{RV} \\
\multicolumn{1}{c|}{}                           & AP   & AP@50   & AP@75   & AP   & AP@50  & AP@75  \\ \hline
RV                                              & -    & -       & -       & 56.70   & 93.40  & 61.70       \\
BEV                                             & 42.70 &84.57 &38.37           & -    & -      & -      \\
Dual-View                                       & \textbf{43.27} &\textbf{84.83} &\textbf{39.76}       & \textbf{57.80}     & \textbf{93.50}       & \textbf{65.10}       \\ \hline
\end{tabular}
\end{small}
\vspace{-6mm}
\end{table}

\begin{figure}
\centering
\noindent
\begin{minipage}[c]{0.46\linewidth}
\centering
    \includegraphics[width=0.99\linewidth]{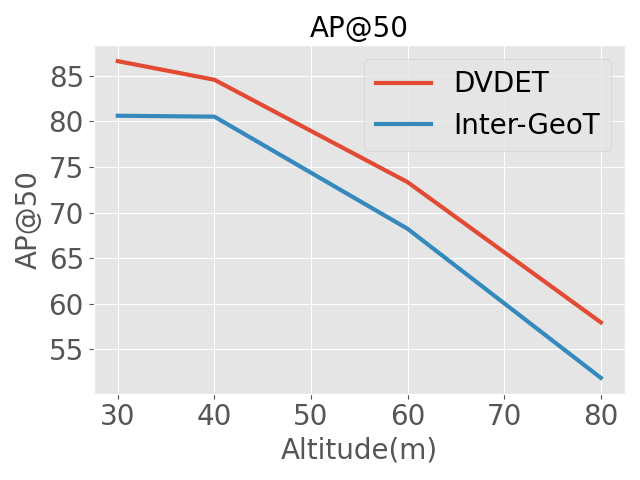}
    \vspace{-6mm}
        \caption{\textit{DVDET} is robust and could alleviate the severe deformation issues at high altitude.}
    \label{fig:AltitudeEffect}
\end{minipage}
\hspace{1mm}
\begin{minipage}[c]{0.455\linewidth}
\centering
    \includegraphics[width=0.99\linewidth]{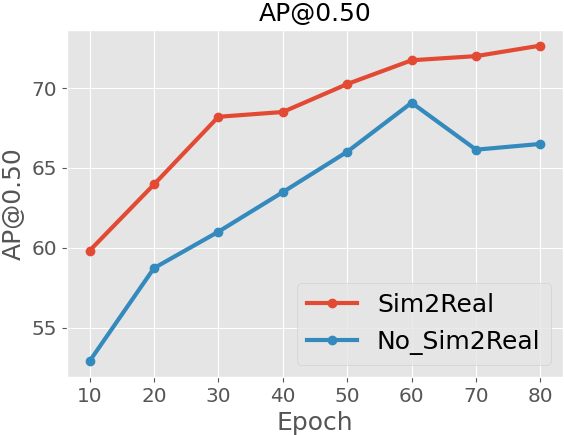}
    \vspace{-6mm}
    \caption{Simulation data benefits BEV object detection on real data with domain transfer.}
    \label{fig:Sim2Real}
\end{minipage}
\vspace{-2mm}
\end{figure}

We next validate the other two proposed modules, 
namely, geo-deformable transformation (\textit{GeoDT}) 
and categorical altitude estimation (\textit{CAE}). Building on \textit{Inter-GeoT}, 
\textit{Inter-GeoDT} integrates~\emph{deformable} convolutions into the geometric transformation and \textit{DVDET} further considers all the possible altitudes.
As shown in Tab.~\ref{tab:MainResults}: 
i) \textit{Inter-GeoDT} consistently performs better than \textit{Inter-GeoT} and improves by 12.99\% on AP@75 on the fullset, reflecting the effectiveness of the proposed geo-deformable transformation;  ii) \textit{Inter-GeoT-CAE} improves \textit{Inter-GeoT} by 7.50\% on AP on the fullset and \textit{DVDET} improves \textit{Inter-GeoDT} by 4.54\%, reflecting the effectiveness of the proposed categorical altitude estimation. 
Fig.~\ref{fig:AltitudeEffect} shows that 
i) BEV detection performance degrades with the altitude, 
which means the detection difficulty is increasing along with the altitude; 
ii) the improvement of \textit{DVDET} over \textit{Inter-GeoDT} is stable across all the altitudes, which means that \textit{DVDET} is robust and alleviates the severe deformation issues at high altitude.

\begin{figure}[!t]
  \centering
  \includegraphics[width=1.0\linewidth]{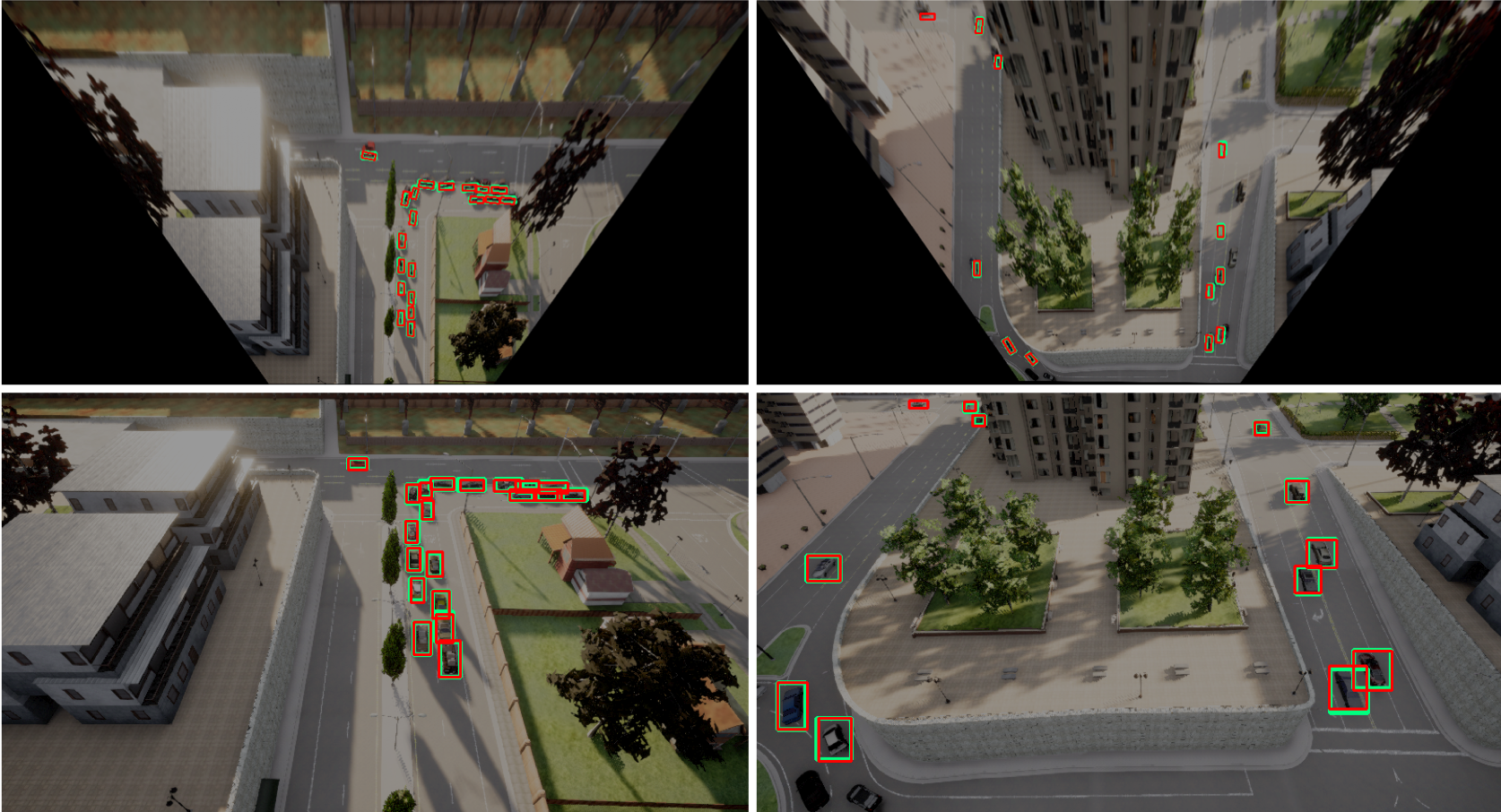}
 \vspace{-4mm}
  \caption{The qualitative results of DVDET on AM3D-Sim. The upper row shows the BEV results and the bottom row shows the RV results. The ground-truth are colored \textcolor{green}{green} and the predictions are colored \textcolor{red}{red}.}
  \label{fig:Simdata_Quantatitive}
  \vspace{-2mm}
\end{figure}

We further validate the feasibility of the dual view object detection in Tab.~\ref{tab:DualView-Sim}. 
We see that: i) dual-view outperforms the individual RV and BEV by 1.1 and 0.57 on AP, respectively. 
It means that the two views can provide complementary information and promote each other: 
BEV can alleviate the occlusion in the RV, 
while RV can provide more object details and more smooth image context to help alleviate the deformation in BEV. 
Fig~\ref{fig:Simdata_Quantatitive} presents quantitative results on AM3D-Sim. We see that: i) the proposed system accurately detects most of the objects in dual-views; ii) the occlusion and overlapping between objects are alleviated in BEV; iii) as the right sample shows, the objects on the ramp can be accurately detected. Note that the presented BEV images are the RV images transformed to the ground plane. 
It only provides a rough idea about BEV object detection and cannot accurately represent the real situation, especially for the areas with varying altitudes. 

\begin{table}[!t]
\vspace{-1mm}
\caption{Overall performance on AM3D-Real. * denotes with domain transfer from simulation to real data.}
\label{tab:DualView-Real}
\vspace{-2mm}
\setlength\tabcolsep{3pt}
\begin{scriptsize}
\begin{tabular}{cl|ccc|ccc}
\hline
\multirow{2}{*}{System} & \multicolumn{1}{c|}{\multirow{2}{*}{Method}}  & \multicolumn{3}{c|}{BEV} & \multicolumn{3}{c}{RV} \\
& \multicolumn{1}{c|}{} & \multicolumn{1}{c}{AP} & \multicolumn{1}{c}{AP@50} & \multicolumn{1}{c|}{AP@75} & \multicolumn{1}{c}{AP} & \multicolumn{1}{c}{AP@50} & \multicolumn{1}{c}{AP@75} \\ \hline
RV                      &-             &-  &- & - & 39.90 & 84.70 & 30.00 \\\hline
\multirow{8}{*}{BEV}    &\textit{Inter-GeoT}      & 22.67	& 62.26  & 9.76  &-  &- & -   \\
                        &\textit{Inter-GeoDT}   & 23.89	& 65.04  & 11.50  &-  &- & -     \\
                        &\textit{Inter-GeoT-CAE}  & 24.13  & 66.41 & 10.71  &-  &- & -   \\
                        &\textit{DVDET} & 26.79  & 69.08 & 13.89  &-  &- & -       \\  
                        &\textit{Inter-GeoT*}      & 25.14 & 60.52  & 17.29  &-  &- & -   \\
                        &\textit{Inter-GeoDT*}   &27.11  &65.95 & 17.05  &-  &- & -     \\
                        &\textit{Inter-GeoT-CAE*}   & 28.19  & 68.79 & 16.30  &-  &- & -     \\
                        &\textit{DVDET*}   &29.04  & \textbf{72.66} & 15.09  &-  &- & -     \\ \hline
\multirow{2}{*}{Dual-View}               
&\textit{DVDET} & 27.39 & 68.46 & 14.43 & 41.60 & 84.80 & 34.10   \\
&\textit{DVDET*} & \textbf{31.82} & 72.60 & \textbf{21.40} & \textbf{43.50} & \textbf{85.90} & \textbf{35.90}   \\ \hline
\end{tabular}
\end{scriptsize}
\vspace{-2mm}
\end{table}

\begin{figure}[!t]
  \centering
  \includegraphics[width=1.0\linewidth]{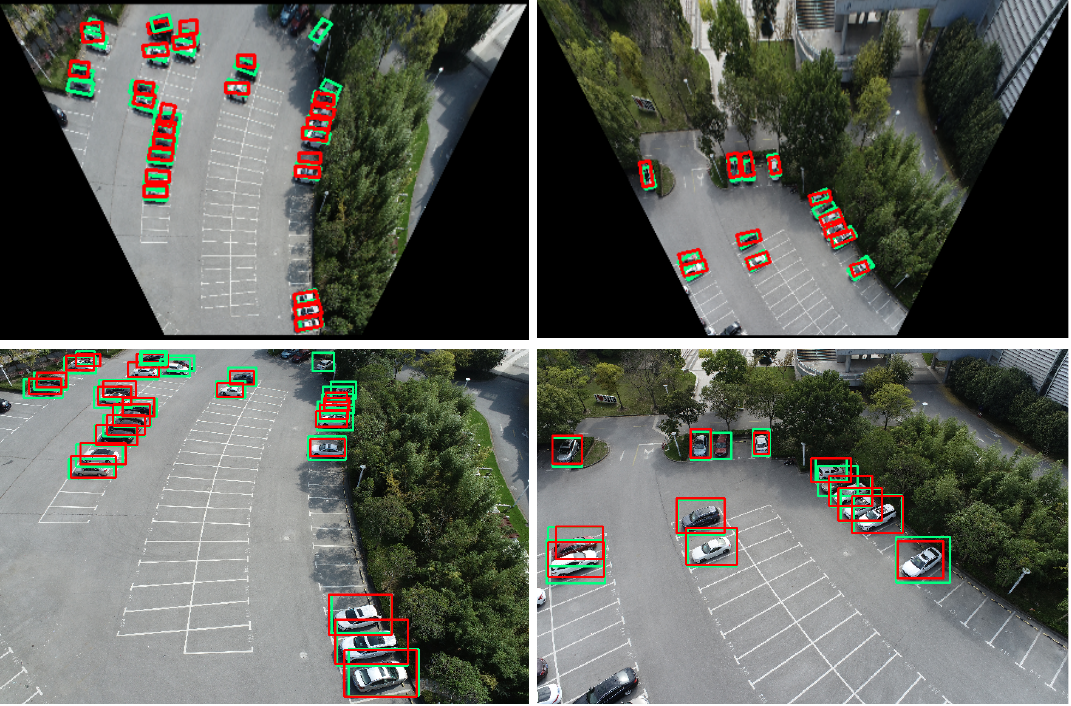}
  \vspace{-4mm}
  \caption{The quantitative results of DVDET on AM3D-Real. The upper row shows the BEV results and the bottom row shows the RV results.}
  \label{fig:Realdata_Qualitative}
  \vspace{-4mm}
\end{figure}

\mypar{Evaluation on AM3D-Real} 
We further validate the proposed modules and dual-view system on the real-world dataset with abundant scenes and high-quality annotations, 
which is relatively smaller than the simulation one due to the costly collection and annotation. 
To mitigate this, we apply the domain adaptation technique to transfer the pre-trained system on the simulation data to handle real-world data. 
From Tab.~\ref{tab:DualView-Real}, we see that: 
i) both the proposed module: geo-deformable transformation (GeoDT), 
and categorical altitude estimation (CAE) achieve improvements, 
the in-line results with cognition validate that the proposed system is robust and the proposed dataset is trustworthy; 
ii) dual-view outperforms the individual RV and BEV; 
iii) the simulation data could effectively help to improve the real-world detection performance. 
Fig.~\ref{fig:Sim2Real} shows that the detector pre-trained on simulation data consistently enables a better initialization for the real-world data. Fig.~\ref{fig:Realdata_Qualitative} presents qualitative results. 
We see: 
i) the proposed system could accurately detect most of the objects in dual-views; ii) the occlusion and overlapping between objects are alleviated in BEV.

\vspace{-2mm}
\subsection{Ablation studies}
\label{sec:ablation}
We provide ablation studies to validate our design choices, and all experiments are conducted on AM3D-Sim.

\mypar{Effect of geo-deformable transformation}
Tab.~\ref{tab:Geo-deformable} explores the properties of geometric and deformable transformation with multiple variants. 
We see that: i) the geometric transformation \textit{Geo} achieves stable but inferior performance; 
ii) the purely learnable transformation without any geometric guidance \textit{MLP} fails; 
iii) the improperly introduced deformable flexibility \textit{Geo-MLP} severely harms the transformation; 
iv) a well-designed deformable module \textit{Geo-DCN} and \textit{Geo-DADCN} boost the detection performance; 
v) the distance prior could help DCN learn the offset, 
and improve the AP from 40.23 to 42.70. 
To sum up, the proposed geo-deformable transformation enjoys both stability of geometric transformation and the flexibility of learnable transformation.

\begin{table}[!t]
\vspace{-1mm}
\small
\caption{Ablation studies on the transformation types. The geo-deformable variant \textit{Geo-DADCN} performs the best.}
\vspace{-2mm}
\centering
\begin{small}
\begin{tabular}{ll|ccc}
\hline
\multicolumn{2}{c|}{Transformation}                              & AP & AP@50 & AP@75 \\ \hline
\multicolumn{1}{l|}{Geometric}                     & \textit{Geo}       &38.64      &80.52      &32.19      \\ \hline
\multicolumn{1}{l|}{Deformable}                      & \textit{MLP}       & 23.56	&58.16	    &14.97      \\ \hline
\multicolumn{1}{l|}{\multirow{3}{*}{Geo-Deformable}} & \textit{Geo-MLP}   & 24.45	&59.16	    &16.26 \\
\multicolumn{1}{l|}{}                                & \textit{Geo-DCN}   & 40.23	&81.83	    &34.83     \\
\multicolumn{1}{l|}{}                                & \textit{Geo-DADCN} & \textbf{42.70}   & \textbf{84.57}   & \textbf{38.37}      \\ \hline
\end{tabular}
\end{small}
\vspace{-2mm}
\label{tab:Geo-deformable}
\end{table}

\begin{table}[!t]
\small
\setlength\tabcolsep{2pt}
\caption{Ablation studies on the categorical altitude estimation. Categorical estimation improves 5.50\% on AP@75.}
\vspace{-2mm}
\label{tab:CategoricalAltitudeEst}
\centering
\begin{footnotesize}
\begin{tabular}{c|ccc|ccc}
\hline
Altitudes       & Estimation  & Supervision & Spacing  & AP  & AP@50  & AP@75  \\ \hline
Single          & -        & -         & -  & 40.84 &81.85 &36.37 \\ \hline
\multirow{4}{*}{Multiple}  & Categorical & - & Prior &40.94	&81.7 &36.85  \\
& Categorical & \checkmark & Uniform &42.08 & 83.62 & 37.48  \\
& Categorical & \checkmark & Prior &\textbf{42.70} &\textbf{84.57} &\textbf{38.37} \\
& Continuous  & \checkmark & - & 40.69  & 83.21 & 35.08   \\ \hline
\end{tabular}
\end{footnotesize}
\vspace{-2mm}
\end{table}


\begin{table}[!t]
\vspace{-1mm}
\caption{\textit{DVDET} outperforms 3D detection SOTAs for cars and 2D detection SOTAs for flat BEV image on aerial 3D detection.}
\label{tab:compare_prev_works}
\vspace{-2mm}
\setlength\tabcolsep{3pt}
\begin{footnotesize}
\begin{tabular}{l|cc|cc|l}
\hline
\multirow{2}{*}{AP {\tiny for method} } & \multicolumn{2}{c|}{ \emph{3D detection for cars}} & \multicolumn{2}{c|}{ \emph{2D detection} } & \multirow{2}{*}{\scriptsize DVDET}     \\
& {\scriptsize MonoRCNN}{\tiny\cite{GDD}}    & {\scriptsize CaDDN}{\tiny\cite{reading2021categorical}}  & {\scriptsize Faster-RCNN}{\tiny\cite{FasterR-CNN}}  & {\scriptsize SwinT}{\tiny\cite{Swin}}      &                           \\ \hline
\multicolumn{1}{c|}{\scriptsize AM3D-Sim} & 0.41   & 41.54   & 20.05 & 24.00   & \multicolumn{1}{c}{\textbf{43.27}} \\ \hline
\multicolumn{1}{c|}{\scriptsize AM3D-Real} & 0.00   & 24.13   & 12.40 & 8.20   & \multicolumn{1}{c}{\textbf{31.82}} \\ \hline
\end{tabular}
\end{footnotesize}
\vspace{-6mm}
\end{table}

\mypar{Effect of categorical altitude estimation}
Tab.~\ref{tab:CategoricalAltitudeEst} assesses the effectiveness of categorical altitude estimation. 
We see that: i) without supervision, the BEV representation with multiple altitudes shows similar performance as its single version.
The minor difference among altitudes is difficult to catch; 
and ii) with the proper altitude guidance, 
the augmented BEV representation across multiple altitudes could achieve superior performance, 
the overall AP could achieve 42.70; 
iii) in the same setting, continuously regressing is clearly worse than categorically binning (40.69 vs. 42.70);
iv) the proposed method is robust to different altitude spacing choices. Uniform spacing achieves comparable performance to our prior-based spacing.

\begin{table}[!t]
\caption{\textit{DVDET} outperforms previous SOTAs by 15.32\% on KITTI, a well-known autonomous driving benchmark.}
\vspace{-4mm}
\begin{center}
\setlength\tabcolsep{6pt}
\begin{tabular}{l | c c }
\hline
\multirow{2}{*}{Method} & \multicolumn{2}{c}{$\text{AP}|_{R_{40}}$~{[Easy~/~Mod~/~Hard~]}~\textuparrow}  \\ 
 & AP$_\text{3D}$  & AP$_\text{BEV}$ \\ \hline
FQNet(CVPR19)~\cite{DBLP:conf/cvpr/LiuLXT019} & 2.77/1.51/1.01 	& 5.40/3.23/2.46 \\ 
M3D-RPN(ICCV19)~\cite{3DDirect4} & 14.76/9.71/7.42 	& 21.02/13.67/10.23 \\ 
MonoPair(CVPR20)~\cite{DBLP:conf/cvpr/ChenTSL20} & 13.04/9.99/8.65 	& 19.28/14.83/12.89 \\ 
MoVi-3D(ECCV20)~\cite{DBLP:conf/eccv/SimonelliBPRK20}  & 15.19/10.90/9.26 	& 22.76/17.03/14.85 \\ 
RTM3D(ECCV20)~\cite{RTM3D} & 14.41/10.34/8.77 	& 19.17/14.20/11.99 \\ 
MonoRCNN(ICCV21)~\cite{GDD} & 18.36/12.65/10.03  & 25.48/18.11/14.10   \\ 
CaDDN(CVPR21)~\cite{reading2021categorical} & 19.17/13.41/11.46 & -/-/-   \\ 
GUP Net(ICCV21)\cite{GUP} & 20.11/14.20/11.77 & -/-/- \\
\textit{DVDET} & \textbf{23.19}/\textbf{15.44}/\textbf{13.07} & \textbf{32.05}/\textbf{22.15}/\textbf{19.32} \\
\hline
\end{tabular}
\end{center}
\vspace{-9mm}
\label{tab:kitti}
\end{table}

\vspace{-3mm}
\subsection{Generalization to autonomous driving}
We further validate the proposed modules by comparing with previous SOTAs on AM3D-Sim and the KITTI 3D object detection benchmark~\cite{geiger2012Kitti}. Tab.~\ref{tab:compare_prev_works} shows \textit{DVDET} clearly outperforms 3D autonomous driving and 2D detection SOTAs on both simulation and real datasets. Note: i) MonoRCNN~\cite{GDD} and CaDDN~\cite{reading2021categorical} are worse than~\textit{DVDET} as they do not consider severe view deformation; while we alleviate this by the proposed geo-deformable transformation; 
ii) Faster-RCNN~\cite{FasterR-CNN} and SwinT~\cite{Swin} on 2D BEV images are worse than~\textit{DVDET} as they consider 3D scenes with flat images; while~\textit{DVDET} considers 3D features with the proposed categorical altitude estimation technique. Tab.~\ref{tab:kitti} shows \textit{DVDET} clearly outperforms previous 3D autonomous driving SOTAs on KITTI. 
Note: i) \textit{DVDET} benefits monocular 3D object detection for autonomous driving; 
ii) our improvement (15.32\%) is significantly superior to previous SOTA GUP Net(ICCV21)\cite{GUP} (4.90\%).

\vspace{-2mm}
\section{Conclusion}
To address the problem of aerial monocular 3D object detection, 
this paper proposes a new dataset, including both simulation (AM3D-Sim) and real-world (AM3D-Real) dataset, as well as a novel monocular 3D object detection system, DVDET, with two core techniques: categorical altitude estimation and geo-deformable transformation. Extensive experiments show that i) DVDET significantly outperforms baseline methods on AM3D-Sim and AM3D-Real, reflecting the effectiveness of the 3D scene understanding from aerial perspective; ii) the model pre-trained on the simulation dataset benefits real-world performance; and iii) DVDET achieves the leading performance on KITTI, reflecting that the proposed method also benefits autonomous driving.


{\small
\bibliographystyle{IEEEtran}
\bibliography{main}
}

\end{document}